\documentclass{article}

\usepackage{microtype}
\usepackage{bbm}
\usepackage{mathtools}
\usepackage{nccmath}
\usepackage{setspace}

\usepackage{caption}
\usepackage{graphicx}
\usepackage{subcaption}

\usepackage{algorithm}

\usepackage{footnote}

\usepackage[linesnumbered,ruled,vlined,algo2e]{algorithm2e}
\SetKwInput{KwInput}{Input}                
\SetKwInput{KwOutput}{Output}              

\SetCommentSty{mycommfont}

\SetAlCapSty{algcapsty}

\usepackage[T1]{fontenc}
\usepackage{wrapfig,lipsum,booktabs}

\usepackage{soul}
\usepackage{dsfont}
\usepackage{enumerate}
\usepackage{enumitem}

\usepackage[normalem]{ulem}

\usepackage{hyperref}
\usepackage{amsmath}
\usepackage{amsthm}
\usepackage{amsfonts}
\usepackage{bbm}
\usepackage{dsfont}
\usepackage[Symbol]{upgreek}
\usepackage{lscape}
\usepackage{balance}
\usepackage{xspace}
\usepackage{float}

\usepackage{enumitem}

\usepackage{wasysym}
\usepackage[table,xcdraw,dvipsnames]{xcolor}
\definecolor{lightgray}{gray}{0.85}
\definecolor{lightergray}{gray}{0.95}

\usepackage{multirow}
\usepackage{array, boldline, rotating}

\usepackage{amssymb}
\usepackage{pifont}
\newcommand{\cmark}{\ding{51}}%

\newcommand{\mc}[1]{\mathcal{#1}}

\renewcommand*\eqref[1]{(\ref{#1})}

\newcommand{\eg}{\emph{e.g.,~}}
\newcommand{\ie}{\emph{i.e.,~}}

\newcommand{\myparagraph}[1]{\vspace{0.07cm}\noindent\textbf{#1}~}

\def\code#1{\texttt{#1}}

\newcommand{\thickhline}{\hlineB{4}}

\definecolor{LightCyan}{rgb}{0.88,1,1}

\usepackage{amsmath,amsfonts,bm}

\def\eqref#1{equation~\ref{#1}}

\def\1{\bm{1}}

\DeclareMathAlphabet{\mathsfit}{\encodingdefault}{\sfdefault}{m}{sl}
\SetMathAlphabet{\mathsfit}{bold}{\encodingdefault}{\sfdefault}{bx}{n}

\newcommand{\alg}{\code{TURN}\xspace}

\newcommand{\mytitle}{Fine-tuning Pre-trained Models for Robustness \\ Under Noisy Labels}
\usepackage[nonatbib, preprint]{neurips_2023}

\usepackage[utf8]{inputenc} 
\usepackage[T1]{fontenc}    
\usepackage{hyperref}       
\usepackage{url}            
\usepackage{booktabs}       
\usepackage{amsfonts}       
\usepackage{nicefrac}       
\usepackage{microtype}      
\usepackage{xcolor}         
\usepackage{microtype}
\usepackage{graphicx}
\usepackage{kotex}

\title{\mytitle}

\author{%
  Sumyeong Ahn\thanks{This work was done while Sumyeong Ahn was at KAIST AI} \\
  Computer Science and Engineering\\
  Michigan State University\\
  East Lansing, MI 48864 \\
  \texttt{sumyeong@msu.edu} \\
  \And
  Sihyeon Kim \\
  KAIST AI \\
  Seoul, South Korea \\
  \texttt{sihk@kaist.ac.kr} \\
  \AND
  Jongwoo Ko \\
  KAIST AI \\
  Seoul, South Korea \\
  \texttt{jongwoo.ko@kaist.ac.kr} \\
  \And
  Se-Young Yun \\
  KAIST AI \\
  Seoul, South Korea \\
  \texttt{yunseyoung@kaist.ac.kr} \\
}

\begin{document}

\maketitle

\begin{abstract}

The presence of noisy labels in a training dataset can significantly impact the performance of machine learning models. To tackle this issue, researchers have explored methods for \emph{Learning with Noisy Labels} to identify clean samples and reduce the influence of noisy labels. However, constraining the influence of a certain portion of the training dataset can result in a reduction in overall generalization performance. To alleviate this, recent studies have considered the careful utilization of noisy labels by leveraging huge computational resources. Therefore, the increasing training cost necessitates a reevaluation of efficiency. In other areas of research, there has been a focus on developing fine-tuning techniques for large pre-trained models that aim to achieve both high generalization performance and efficiency. However, these methods have mainly concentrated on clean datasets, and there has been limited exploration of the noisy label scenario. In this research, our aim is to find an appropriate way to fine-tune pre-trained models for noisy labeled datasets. To achieve this goal, we investigate the characteristics of pre-trained models when they encounter noisy datasets. Through empirical analysis, we introduce a novel algorithm called \alg, which robustly and efficiently transfers the prior knowledge of pre-trained models. The algorithm consists of two main steps: (1) independently tuning the linear classifier to protect the feature extractor from being distorted by noisy labels, and (2) reducing the noisy label ratio and fine-tuning the entire model based on the noise-reduced dataset to adapt it to the target dataset. The proposed algorithm has been extensively tested and demonstrates efficient yet improved denoising performance on various benchmarks compared to previous methods.

\end{abstract}

\section{Introduction}
\label{sec:intro}

Deep neural networks (DNNs) have demonstrated remarkable performance in various tasks, including classification~\cite{he2016deep}, generation~\cite{goodfellow2020generative}, and object detection~\cite{he2017mask}. However, the effectiveness of DNNs declines significantly when trained with corrupted annotations. Additionally, manually correcting noisy labels or acquiring clean labels anew is challenging due to the large-scale nature of datasets. To tackle this issue, researchers have developed diverse approaches within the field of \emph{Learning with Noisy Labels} (LNL). These approaches encompass robust training loss~\cite{zhang2018generalized, wang2019symmetric}, regularization~\cite{cao2020heteroskedastic, cheng2023mitigating, ko2022alasca}, sample selection~\cite{han2018co, yu2019does}, and semi-supervised learning methods~\cite{li2020dividemix, liu2020early, karim2022unicon}.

However, recent studies in the field of LNL have shown an increase in computational complexity. As demonstrated in~\autoref{tab:time_scrach}, a recent algorithm called UNICON~\cite{karim2022unicon} incurs a substantial training time increase of $622\%$ compared to the vanilla approach. This is primarily due to the utilization of both clean labeled data and noisy labeled samples. Notably, the most computationally expensive aspect is the careful integration of noisy labeled samples. For instance, UNICON employs an expensive contrastive loss that does not rely on labels, resulting in higher computational costs. The motivation behind such costly incorporation of noisy labels is to enhance the generalization performance, which can be derived from a larger training dataset.

We are motivated by the conjecture that pre-trained models (PTMs), widely recognized for their strong generalization performance with fast adaptation, have the potential to effectively address the limitations of previous denoising algorithms. Recent research supports this intuition by demonstrating that PTMs can quickly adapt to new target datasets using few-shot correctly labeled samples and achieve exceptional generalization performance across a wide range of tasks~\cite{devlin-etal-2019-bert, brown2020language, he2022masked, assran2022masked}. This ability is attributed to their robust feature extraction capabilities. Therefore, leveraging PTMs in LNL methods can contribute to improving generalization performance within a few epochs while simultaneously reducing computational costs by minimizing the involvement of potentially noisy labels.

However, there has been limited research on the application of PTMs to noisy labeled datasets and the effective utilization of their valuable knowledge in such scenarios. Therefore, there is a need to investigate methods that are both efficient and effective in leveraging the robust feature extractor of PTMs in the presence of noisy labels.

\begin{figure*}[t]
    \centering
    \includegraphics[width=0.95\textwidth]{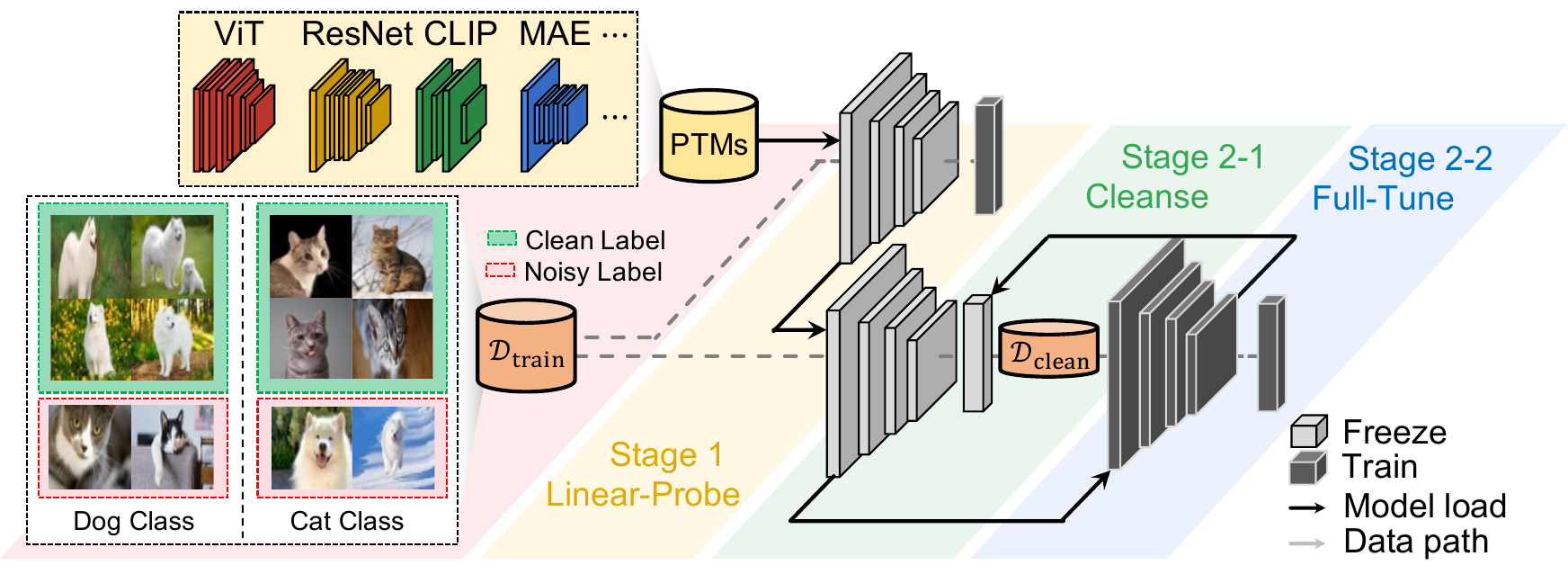}
    \vspace{-5pt}
    \caption{Illustration of the proposed algorithm. We leverage pre-trained models to tune the model by using the noisy-labeled dataset. The proposed algorithm consists of two main parts: (1) Linear probing: update linear classifier only. (2) Iteratively cleansing the given dataset and Full fine-tuning: update the entire parameter based on the cleansed subset of the given noisy dataset.}
    \label{fig:intro}
    \vspace{-5pt}
\end{figure*}
\begin{table}[t!]
\centering
\resizebox{0.98\textwidth}{!}{
\begin{tabular}{p{0.15\textwidth}|>
                {\centering}p{0.13\textwidth}>
                {\centering}p{0.13\textwidth}>
                {\centering}p{0.13\textwidth}>
                {\centering}p{0.13\textwidth}>
                {\centering}p{0.13\textwidth}>
                {\centering\arraybackslash}p{0.13\textwidth}}
\thickhline
                Algorithm
                & Vanilla & GCE~\cite{zhang2018generalized} 
                & ELR~\cite{liu2020early} 
                & ELR+~\cite{liu2020early} 
                & DivideMix~\cite{li2020dividemix} 
                & UNICON~\cite{karim2022unicon}\\ \hline
Cost (Min.)         &108.6     & 110.7 & 118.9 & 248.8 & 574.1 & 675.6 \\
\thickhline
\end{tabular}}
\caption{Training time on CIFAR-100 dataset with symmetric $60\%$ noise case from scratch. }
\label{tab:time_scrach}
\vspace{-20pt}
\end{table}

\myparagraph{Contribution.} This research introduces a robust method for transferring the knowledge of PTMs to a target dataset that probably has noisy labels. To this aim, two adaptation methods are explored: full fine-tuning (FFT) and linear probing (LP). In the FFT approach, all parameters of the PTM are updated, while in the LP tune only the last fully connected (FC) layer with frozen feature extractor. The main observations and contributions of this research are summarized as follows:

\begin{itemize}[leftmargin=*]
\item It is confirmed that when a high proportion of noisy labels is present, the feature extractor becomes distorted when tuned using FFT. Conversely, an improvement in the feature extractor, extracting class-wise good features, is observed when the severity of noisy labels is low.\footnote{A similar observation was noted in a previous study~\cite{cheng2023mitigating}, but their findings were only focused on a SimCLR model trained on the target dataset. However, we expand this to the models trained on different datasets (\eg ImageNet) and various PTMs.}
\item The proposed method, fine-\textbf{TU}ning  pre-trained models for \textbf{R}obustness under \textbf{N}oisy labels  (\alg), consists of two adaptation steps: LP and FFT. The LP step aims to adapt the model to the target task without compromising the integrity of the feature extractor, enabling the effective detection of noisy samples. The FFT step further enhances the feature extractor by updating entire trainable parameters, followed by dataset cleansing. An overview of \alg is presented in Figure \ref{fig:intro}.
\item Experimental results demonstrate the efficiency and robustness of the proposed method compared to existing LNL methods on various datasets, including synthetically noisy-labeled datasets like CIFAR-100, as well as real-world datasets such as WebVision and Clothing1M.
\end{itemize}

\section{Preliminary}
\label{sec:pre}
In this section, we present the necessary background information on two key topics: LNL and PTMs.

\subsection{Learning with Noisy Label} 
Let us denote the training dataset as $\mathcal{D}_{\text{train}} = {(x_i, \bar{y}_i)}_{i=1}^{N}$, comprising $N$ pairs of input images $x_i$ and their corresponding labels $\bar{y}_i \in \{1,\ldots,C\}$. In real-world scenarios, the given label $\bar{y}_{i}$ can be corrupted due to various factors, such as human errors in crowdsourced labeling systems. We use $y_i$ to denote the ground-truth label for $(x_i, \bar{y}_i)$, which is not accessible during the training phase. It is widely acknowledged that models trained on noisy labeled datasets $\mathcal{D}_{\text{train}}$ using conventional classification loss functions, such as cross-entropy loss ($\mathcal{L}_{\text{CE}}$), often exhibit poor performance on test data. This limitation poses challenges when deploying such models in real-world scenarios. Therefore, training robust models capable of effectively handling noisy labeled datasets becomes essential.

\subsection{Pre-trained Model and Fine-tuning}
In recent times, several PTMs have been proposed for image-related tasks, encompassing models trained using supervised learning (SL)~\cite{dosovitskiy2021an, liu2021swin, liu2022convnet}, self-supervised learning (SSL)\cite{he2022masked, assran2022masked}, and multimodal learning~\cite{radford2021learning} have become accessible and demonstrated significant performance in classification tasks. To harness the power of the pre-trained feature extractor, two primary tuning methods are commonly employed: linear probing (LP) and full fine-tuning (FFT). We represent the classification model with a PTM as $g(f(x;\theta);\phi)$, where $g(\cdot; \phi)$ denotes the linear classifier with its parameter $\phi$, and the feature extractor $f(\cdot; \theta)$ has its parameter $\theta$. In LP, $\theta$ remains frozen, and only $\phi$ is trained. Conversely, FFT involves tuning all the trainable parameters, including both $\theta$ and $\phi$. The advantages and disadvantages of LP and FFT are fundamentally distinct. LP has the strength of inexpensive computational cost thanks to the limited number of parameters, but it has restricted adaptability. On the other hand, FFT has strong adaptability, but it requires a huge amount of training resources.

\section{Fine-tuning Pre-trained Models under Label Noise} \label{sec:motivation}

In this section, our objective is to analyze the behavior of the feature extractor when tuned on severely or slightly noisy labels. We particularly focus on identifying the conditions that lead to improvements in the feature extractor. The key observations that form the basis of our investigation are summarized as follows:
\vspace{-7pt}
\begin{enumerate}[leftmargin=1.2cm, label=\textbf{Obs~\arabic*}.]
\item The presence of a high proportion of noisy labels can significantly distort the feature extractor when FFT is applied.
\item Conversely, when the noise ratio is not severe, FFT can effectively enhance the feature extractor, allowing it to construct class-wise clusters accurately.
\end{enumerate}
\vspace{-7pt}
Inspired by these observations, we propose a denoising algorithm for PTMs, which is elaborated in Section~\ref{sec:method}. In the subsequent paragraphs, we provide detailed explanations for these two motivations.

\myparagraph{Experimental Setting.} 
In our investigation, we conduct experiments to evaluate the performance of several popular PTMs on a dataset with noisy labels. The considered PTMs include ViT-B/16~\cite{dosovitskiy2021an}, ConvNeXt-T~\cite{liu2022convnet}, CLIP-ViT-B~\cite{radford2021learning}, and MAE-ViT-B~\cite{he2022masked}. To generate the noisy labeled dataset, we employ the symmetric label-flipping technique on the CIFAR-100 dataset. This technique randomly flips the labels of either $90\%$ or $10\%$ of the samples, resulting in different levels of label noise. We investigate the impact of noisy labels by visualizing the embedding of randomly sampled ten classes from the test dataset using t-SNE~\cite{van2008visualizing} plots. Each model is trained for $5$ epochs, following the convention in previous works~\cite{kumar2022fine}.


\begin{figure*}[t]
    \centering
    \includegraphics[width=1.\linewidth]{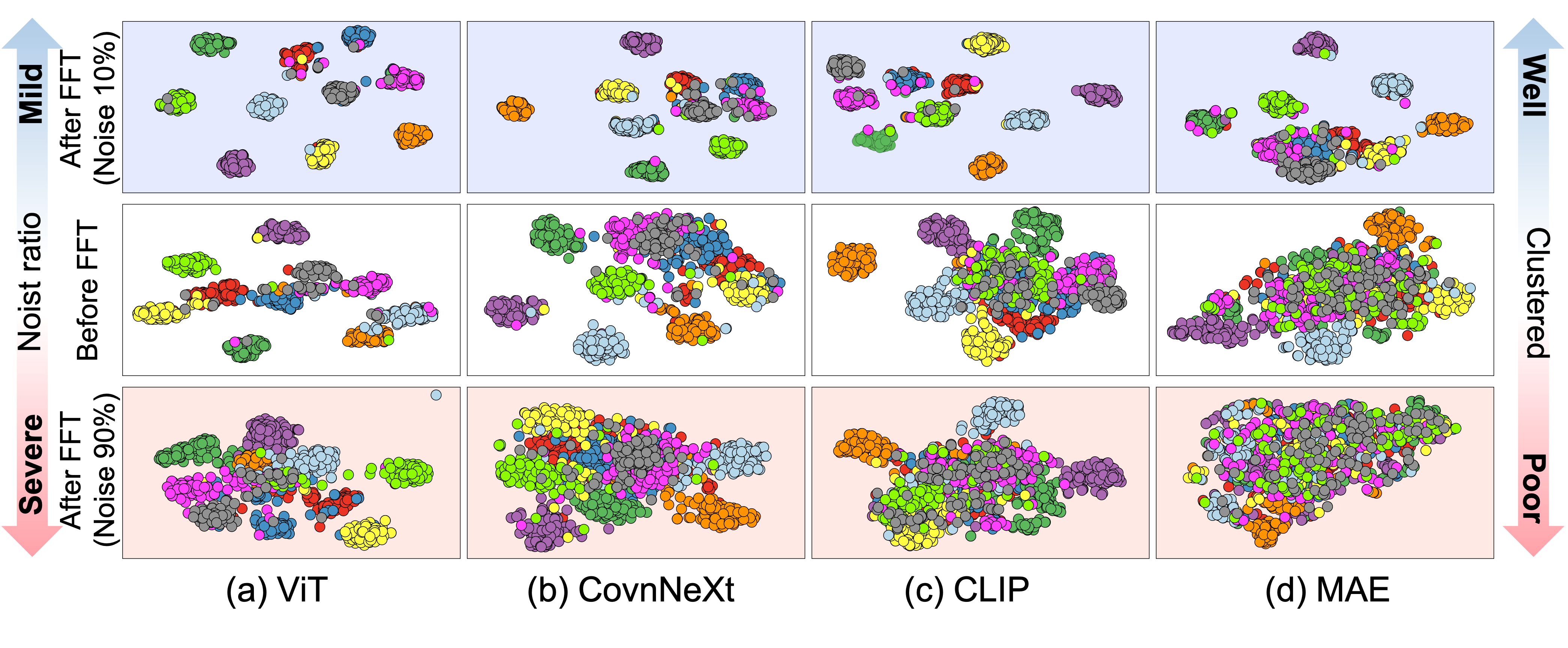}
    \vspace{-23pt}
    \caption{Illustration about tuning characteristics under noisy labeled dataset. We plot t-SNE results before and after FFT on the noise ratio of $90\%$ and $10\%$ datasets. Simply speaking, $60\%$ shows well-clustered features while $90\%$ shows poorly-clustered result.}
    \label{fig:motivation}
    \vspace{-17pt}
\end{figure*}

\myparagraph{(1) In the severe noise case, applying FFT results in distortion of the feature extractor.} 
The distortion of the feature extractor occurs due to the incorrect supervision signals by the noisy labels. To gain a deeper understanding of this phenomenon, we employ t-SNE plots. In Figure~\ref{fig:motivation}, the middle row illustrates the initial feature extractor of each PTM, while the bottom row displays the features after applying FFT with $90\%$ noisy labels. It can be observed that the features become mixed and less distinguishable. This effect is particularly evident when the initial clusters of the PTMs are already poorly defined, \ie MAE. As a result, applying FFT on a severely noisy dataset leads to a trained feature extractor that fails to adequately capture the intrinsic characteristics of the samples.

\myparagraph{(2) FFT improves the feature extractor when the target dataset has slight label noise.} When dealing with minor label noise in the target dataset, applying FFT to the feature extractor yields improvements to construct class-wise clusters effectively. The top row of Figure~\ref{fig:motivation} reveals that the models under examination exhibit better construction of class-wise clusters compared to the middle row. This observation suggests that fine-tuning the feature extractor on a dataset with slight label noise can enhance its performance, enabling it to capture more meaningful features from each sample.

\myparagraph{Key insights.} Based on our analysis of extracted features before and after applying FFT on severely or slightly noisy datasets, we have derived the following insights. When dealing with a slightly noisy dataset, fine-tuning the entire trainable parameters, including FFT, enhances the ability of the feature extractor to cluster samples within each class. It leverages the proficiency of the initial feature extractor and adapts it more specifically to the target task. However, in the case of severe label noise, it is crucial to protect the feature extractor from distortion, as applying FFT can degrade its performance due to the provision of incorrect supervision by the noisy labels. These insights emphasize the importance of considering the severity of label noise to develop an algorithm by leveraging the precious feature extractor without ruining it. Because the noise ratio is unknown, it is important to develop an algorithm that carefully utilizes the feature extractor.

\section{Fine-tuning Pre-trained Model for Learning with Noisy Labels} 
\label{sec:method}

Based on the findings in Section~\ref{sec:motivation}, we can extract crucial strategies for the effective use of PTMs in noisy label datasets, even when the noise ratio is unknown. Firstly, if the training dataset is significantly tainted with label noise (\ie \textbf{Obs 1}), it is critical to avoid directly incorporating these noisy labels as they can negatively impact the valuable feature extractor of PTMs. Hence, a buffer, such as LP, is necessary to manage these highly noisy instances, as it allows the feature extractor to remain frozen. Secondly, when the training dataset exhibits relatively low label noise (\ie \textbf{Obs 2}), FFT should be employed for task adaptation. This is due to the need for the feature extractor to be adjusted to the target dataset after FFT in the presence of minor noise. However, a key challenge remains: the absence of specific information about the noise label, including the intensity of the noise ratio. Therefore, in the absence of noise label-related details, the suggested approach to protect the feature extractor is a two-step procedure involving LP, followed by FFT. This section introduces our proposed two-step method, referred to as \alg (fine-\textbf{TU}ning pre-trained models for \textbf{R}obustness under \textbf{N}oisy labels), aimed at effectively addressing these challenges.


\subsection{Algorithm Description} 
The algorithm consists of two main steps. In Step 1, the algorithm utilizes LP with an initialized fully connected layer, denoted as $g(z;\phi)$, where $z$ represents the output of the frozen feature extractor $f(x;\theta)$ for each input image $x$. Subsequently, in Step 2, the algorithm iteratively cleanses the training dataset and performs FFT on the entire model $g(f(x;\theta);\phi)$ using a subsampled dataset $\mathcal{D}_{\text{clean}}$. The complete procedure of the proposed algorithm is described in Algorithm~\ref{alg:alg}.

\begin{algorithm}[t!]
    \DontPrintSemicolon
    \SetAlgoLined
    \SetNoFillComment
    \LinesNotNumbered
    \caption{Pseudo code of \alg}
    \label{alg:alg}
    \KwInput{Dataset $\mc{D}_{\text{train}}=\{(x_i, \bar{y}_i)\}_{i=1}^{N}$, Linear classifiers $g(z;\phi)$, Pre-trained feature extractor $f(x;\theta)$, GMM threshold $\tau$, Linear probing epoch $E_{\text{LP}}$, Full fine tuning epoch $E_{\text{FFT}}$}
    \KwOutput{Fine-tuned model $g(f(\cdot))$}
    \raggedright
    \textcolor{gray}{/* Step 1: Linear probing */} \\
    Extract feature $z_i = f(x_i;\theta)$ for all samples $(x_i, \bar{y}_i) \in \mc{D}_{\text{train}}$\\
    \For{$e < E_{\text{LP}}$}
    {
        Train the linear classifier $g(z_i;\phi)$ under GCE Loss: $\mc{L}_{\text{GCE}}(g(z_i;\phi), \bar{y}_i)$ \\
    }
    \textcolor{gray}{/* Step 2: Select clean samples and FFT */}\\
    \For{$e < E_{\text{FFT}}$}
    {
        Extract per-sample loss for all samples in $\mc{D}_{\text{train}}$ \\
        Construct clean subset $\mc{D}_{\text{clean}}$ by using GMM with threshold $\tau$\\
        Train the model $g(f(x_j;\theta);\phi)$ under  $\mc{L}_{\text{CE}}(g(f(x_j;\theta)\phi),  \bar{y}_j)$ on $\mc{D}_{\text{clean}} = \{(x_j, \bar{y}_j)\}$ 
    }
\end{algorithm}

\subsubsection{Step 1: Linear Probing}
The first step of the algorithm serves two main objectives. Firstly, it aims to obtain a classifier that can effectively detect noisy labels, which aligns with the approach taken by previous works that utilize a warm-up phase before detecting noisy labels~\cite{li2020dividemix, karim2022unicon, kim2021fine}. Secondly, during the training of the classifier for detecting noisy labels, it is crucial to ensure that the feature extractor is not distorted by the noisy labels. To address this, we employ LP, which helps protect the feature extractor from being affected by severe label noise. As mentioned in Section~\ref{sec:motivation}, there is evidence that the feature extractor can be improved through FFT when applied to datasets with a small proportion of noisy labels. However, since the exact degree of label noise is unknown, we adopt a conservative approach by freezing the feature extractor until we have a classifier capable of detecting noisy labels. 

Additionally, applying LP enhances efficiency, which is a key consideration derived from the inherent strengths of LP, as described in Section~\ref{sec:pre}. Therefore, we employ the following implementation technique. Since the feature extractor remains frozen throughout the algorithm, the extracted features of input images remain unchanged. As a result, we pre-extract the features $z_i$ of input images $x_i$ as $z_i = f(x_i;\theta)$ in advance, minimizing computational overhead during the subsequent steps.
\begin{equation*}
    \mc{Z} = \left \{z_i | f(x_i;\theta)  \right \} \quad \forall (x_i, \bar{y}_i) \in \mc{D}_{\text{train}}.
\end{equation*}
Next, we update the parameters $\phi$ for $E_{\text{LP}}$ epochs by utilizing the Generalized Cross Entropy loss~\cite{zhang2018generalized}, denoted as $\mathcal{L}_{\text{GCE}} = \frac{1 - g(z_i; \phi)^{q}}{q}$, where $q$ is a hyperparameter. The GCE loss is employed to mitigate the influence of noisy labels during the training of the linear classifier. This step is crucial for effectively training the linear classifier, as it helps cleanse the dataset in preparation for the subsequent steps.

\subsubsection{Step 2: Cleansing and FFT} In Step 2, the main objective is to improve the performance of the model by adapting it to the target dataset. As mentioned in Section~\ref{sec:motivation}, applying FFT to a dataset with a slight noise label enables the feature extractor to better adapt to the target dataset. Therefore, the remaining part of this step focuses on obtaining a sufficiently cleansed dataset. This stage is divided into two sub-steps: cleansing and FFT. In the cleansing phase, clean samples are selected from the given noisy dataset, while discarding the remaining samples. Following the cleansing phase, FFT is carried out on the selected clean subset, to improve the performance of the model on the target dataset. These two sub-steps are conducted for $E_{\text{FFT}}$ epochs, iteratively. This is because a model that is trained better exhibits a greater ability to distinguish noisy labels.

Note that we can achieve efficiency by discarding the noisy labeled samples during the FFT procedure, leveraging the inherent strength of PTMs. As mentioned earlier, PTMs are highly capable of adapting to clean datasets using few-shot learning techniques. This allows the model to achieve successful performance even with a small portion of clean training samples. In the subsequent sections, we will provide a comprehensive description of each sub-step involved in the FFT process.

\myparagraph{Step 2-1: Selecting clean samples.} 
To identify clean samples, we utilize a clustering algorithm based on the Gaussian Mixture Model (GMM), a commonly employed in previous research works~\cite{li2020dividemix, kim2021fine, ahn2022denoising}. This algorithm begins by calculating the loss for each sample, which is determined based on its noisy labels. This process allows us to distinguish between clean and noisy samples in the dataset.
\begin{equation*}
    \ell_i = \mc{L}_{\text{CE}} (g(f(x_i;\theta);\phi), \bar{y}_i) \quad \text{ where } \quad (x_i, \bar{y}_i) \in \mc{D}_{\text{train}}.
\end{equation*}
After obtaining the per-sample loss, the next step involves fitting the GMM model using the calculated losses, denoted as $\ell_i$. From this GMM model, we extract two Gaussian distributions for each class~$c$: $p_{\text{l}}^{c}$ and $p_{\text{h}}^{c}$. Here, $p_{\text{l}}^{c}$ represents the distribution with a lower mean value compared to $p_{\text{h}}^{c}$. Utilizing these per-class distributions, we construct the clean dataset using the following procedure:
\begin{equation*} 
    \mc{D}_{\text{clean}} = \bigcup_{c=1}^{C} \mc{U}(\mc{D}_{\text{clean}}^{c}, N) \quad \text{ where } \quad \mc{D}_{\text{clean}}^{c} = \{(x_i, \bar{y}_i) | p_{l}^{c} (\ell_i) > \tau \text{ where } \bar{y}_i = c \}.
\end{equation*} 
In this formulation, the threshold hyperparameter $\tau$ is used, and the function $\mc{U}(\mc{D},N)$ denotes the uniform sampling function, which randomly selects $N$ samples uniformly from the set $\mc{D}$. It is worth noting that an equal number of samples are selected for each class from the set considered clean. This selection strategy is based on the principle emphasized in the work~\cite{karim2022unicon}, which highlights the importance of maintaining a uniform distribution. Therefore, we set $N = \min_{c\in\{1,...,C\}} |\mc{D}_{\text{clean}}^{c}|$, where $|\mc{D}|$ represents the cardinality of the set $\mc{D}$.

\myparagraph{Step 2-2: FFT on the selected dataset $\mc{D}_{\text{clean}}$.} In Step 2-1 of the \alg, a clean subset is extracted from the noisy target dataset. This subset is obtained by reducing the proportion of noisy labels, thereby creating an environment conducive to conducting FFT. With a reduced influence from noisy labels, the FFT process can be executed, allowing the model to adapt more effectively to the underlying clean samples present in the training dataset.

\section{Experiments}
\label{sec:exp}

In this section, we present empirical evaluations that showcase the superior performance and computational efficiency of our proposed algorithm in handling noisy labels. We begin by providing a detailed description of the LNL benchmarks and implementation in Section~\ref{sec:exp_setting}. Subsequently, in Section~\ref{sec:exp_result}, we present the experimental results obtained from extensive evaluations. Additionally, we conduct additional experiments to gain a deeper understanding of \alg in Section~\ref{sec:exp_analysis}.

\subsection{Experimental Setting}
\label{sec:exp_setting}

\myparagraph{Datasets.} To begin our evaluation, we assess the performance of \alg on widely-used image classification tasks that involve noisy labeled datasets. Specifically, we conduct an evaluation on the synthetically noised CIFAR-100 dataset, and real-world noisy labeled dataset, the Clothing 1M and WebVision datasets. For the CIFAR dataset, we introduce uniform random noise into a portion of the labels to simulate symmetric noise, asymmetric noise which flips the labels to specific classes~\cite{liu2020early}. To incorporate instance-dependent noise, we adopt the noise generation methodology described in~\cite{cheng2021learning}.

\myparagraph{Architectures and baselines.} In our evaluation, we consider several PTMs for each dataset, including ViT-B/16~\cite{dosovitskiy2021an}, ConvNeXt-T~\cite{liu2022convnet}, MAE-ViT-B~\cite{he2022masked}, MSN-ViT-B~\cite{assran2022masked}, CLIP-ViT-B~\cite{radford2021learning}, and ResNet-50~\cite{he2016deep}. In the case of CLIP model, we utilize the visual model. Each weight of PTMs is obtained from HuggingFace~\cite{wolf-etal-2020-transformers}. Among these architectures, we compare our proposed \alg with various previous methods, including Vanilla~(trained on cross-entropy loss), GCE~\cite{zhang2018generalized}, ELR~\cite{liu2020early}, DivideMix~\cite{li2020dividemix}, and UNICON~\cite{karim2022unicon}. Additionally, we apply both FFT and LP to all algorithms for comparison. However, it is worth noting that DivideMix and UNICON require a significant number of feed-forwards for each sample, approximately 4 times and 8 times, respectively, with various data augmentation techniques. Due to the computational limitation, we utilize LP for both DivideMix and UNICON. For further details on the implementation, please refer to Appendix~\ref{app:base}.

\myparagraph{Implementation.} To optimize the hyperparameters for each model, we utilize the \code{Ray}~\cite{liaw2018tune} hyperparameter tuning tool. This allows us to identify the appropriate settings for parameters such as learning rate, weight decay, optimizer, and batch size. For each PTM, specific optimized hyperparameters and their search spaces are described in the Appendix~\ref{app:imp}. Regarding the hyperparameter for \alg, namely GMM threshold, $\tau=0.6$, we adopt the value suggested in the DivideMix paper~\cite{li2020dividemix}, which initially introduced the GMM-based clean sample selection. For baselines, we run $5$ epochs for FFT and $20$ epochs for LP, while \alg is optimized $20$ epochs of LP with $4$ epochs for FFT in Step 2 to spend smaller computational cost compared to the baselines.

\begin{table*}[t]
\centering
\caption{Comparison with LNL algorithms in test accuracy (\%) on CIFAR-100 dataset with symmetric, asymmetric, and isntance noise. We run totally six architectures under the same noisy label setting. The best results are highlited in \textbf{bold}. We report the best/last performance for each experiment.}
\label{tab:cifar_synth}
\vspace{-5pt}
\resizebox{0.98\textwidth}{!}{
\begin{tabular}{cc|cccc|cccc}
\thickhline
\multirow{2}{1cm}{Tuning \\ \,\, Type}    & \multirow{2}{1cm}{Method}  
                        & \multicolumn{8}{c}{CIFAR-100}\\ 
                        
                        && \textit{Symm.} $0.6$ & \textit{Symm.} $0.9$ & \textit{Asym.} $0.4$ & \textit{Inst.} $0.4$ 
                        & \textit{Symm.} $0.6$ & \textit{Symm.} $0.9$ & \textit{Asym.} $0.4$ & \textit{Inst.} $0.4$ \\\hline \hline
&& \multicolumn{4}{c|}{ViT-B/16} & \multicolumn{4}{c}{ConvNeXt-T} \\ \hline
& CE 
                        & 88.56 / 77.39 & 63.43 / 49.61 & 61.94 / 61.45 & 64.82 / 62.07
                        & 79.38 / 75.06 & 54.15 / 49.39 & 68.72 / 59.98 & 57.81 / 55.34 \\
FFT & GCE 
                        & 89.98 / 89.98 & 45.40 / 45.40 & 83.09 / 83.86 & \,1.61 / \,1.61 
                        & 82.13 / 82.25 & 64.06 / 64.06 & 79.87 / 72.01 & \,1.07 / \,1.07 \\
& ELR 
                        & 89.03 / 84.15 & 64.12 / 58.05 & 77.36 / 74.65 & 83.04 / 71.40
                        & 79.24 / 77.21 & 51.71 / 48.01 & 74.94 / 73.03 & 66.76 / 52.95\\ 
\rowcolor{lightergray}
& CE 
                        & 81.00 / 74.70 & 64.61 / 39.50 & 61.66 / 57.18 & 61.66 / 57.81 
                        & 70.30 / 64.54 & 52.52 / 33.40 & 54.08 / 48.62 & 61.99 / 52.97 \\
\rowcolor{lightergray}
& GCE 
                        & 84.63 / 84.63 & 80.92 / 80.92 & 76.05 / 76.05 & 43.21 / 43.21 
                        & 74.74 / 74.74 & 65.67 / 65.67 & 70.91 / 70.91 & \,5.17 / \,5.17 \\
\rowcolor{lightergray}
LP & ELR 
                        & 81.03 / 75.88 & 65.80 / 42.43 & 64.50 / 57.09 & 69.42 / 66.30 
                        & 70.90 / 65.58 & 52.31 / 33.88 & 56.47 / 53.49 & 61.40 / 56.87 \\
\rowcolor{lightergray}
& DivideMix 
                        & 84.28 / 84.23 & 80.77 / 80.45 & 83.00 / 83.00 & 84.55 / 84.50  
                        & 75.59 / 75.59 & 68.10 / 67.85 & 72.79 / 72.79 & 65.51 / 65.51 \\
\rowcolor{lightergray}
& UNICON 
                        & 84.25 / 84.10 & 80.84 / 79.82 & 84.18 / 84.10 & 84.87 / 84.69 
                        & 70.25 / 70.25 & 60.11 / 60.11 & 65.27 / 65.24 & 69.25 / 69.25 \\ 

\rowcolor{lightgray}
LP-FFT & Ours
                        & \textbf{90.50 / 90.50} & \textbf{84.53 / 83.33} & \textbf{87.33 / 87.33} & \textbf{87.77 / 87.77}
                        & \textbf{83.74 / 83.74} & \textbf{69.94 / 69.94} & \textbf{81.68 / 81.37} & \textbf{73.10 / 73.10} \\ \hline \hline
&& \multicolumn{4}{c|}{MAE-ViT-B} & \multicolumn{4}{c}{MSN-ViT-B} \\ \hline
& CE 
                        & 60.30 / 61.68 & \,8.00 / \,8.00 & 55.80 / 55.03 & 50.41 / 50.06
                        & 67.11 / 67.11 & \,5.88 / \,5.88 & 57.61 / 58.95 & 62.70 / 51.69 \\
FFT & GCE 
                        & 58.30 / 58.30 & \,3.16 / \,3.16 & 60.03 / 60.03 & \,1.00 / \,1.00 
                        & 65.03 / 65.03 & \,7.71 / \,7.71 & 61.58 / 61.58 & \,1.00 / \,1.00 \\
& ELR 
                        & 63.10 / 63.10 & \,7.96 / \,7.96 & 66.83 / 66.83 & 48.68 / 48.68
                        & 67.25 / 67.25 & \,5.90 / \,5.90 & 70.15 / 65.51 & 57.93 / 57.93\\ 
\rowcolor{lightergray}
& CE 
                        & 48.11 / 47.08 & 19.92 / 13.41 & 38.89 / 36.56 & 44.64 / 42.41 
                        & 59.18 / 56.65 & 22.28 / 21.73 & 47.92 / 47.05 & 63.99 / 40.99 \\
\rowcolor{lightergray}
& GCE 
                        & 49.72 / 49.36 & 14.40 / 14.02 & 48.72 / 47.12 & \,1.69 / \,1.69 
                        & 47.57 / 47.57 & 14.54 / 14.54 & 42.74 / 42.52 & \,1.75 / \,1.75 \\
\rowcolor{lightergray}
LP & ELR 
                        & 47.66 / 46.86 & 17.26 / 13.73 & 39.28 / 37.70 & 46.03 / 38.87 
                        & 60.18 / 56.62 & 20.20 / 18.84 & 50.59 / 46.08 & 61.81 / 54.23 \\
\rowcolor{lightergray}
& DivideMix 
                        & 59.99 / 59.84 & 24.27 / 23.22 & 55.45 / 55.29 & 50.98 / 50.98 
                        & 70.88 / 70.49 & 42.98 / 42.31 & 65.65 / 65.08 & 61.36 / 60.54 \\
\rowcolor{lightergray}
& UNICON 
                        & 37.57 / 38.57 & 21.45 / 21.44 & 34.06 / 34.06 & 39.05 / 39.05 
                        & 67.36 / 67.36 & 51.59 / 51.46 & 60.94 / 60.94 & 66.13 / 66.13 \\ 

\rowcolor{lightgray}
LP-FFT & Ours
                        & \textbf{64.63 / 64.63} & \textbf{28.52 / 28.52} & \textbf{65.72 / 65.72} & \textbf{56.43 / 56.43}
                        & \textbf{79.42 / 79.42} & \textbf{54.53 / 54.53} & \textbf{75.11 / 74.95} & \textbf{68.91 / 68.91}\footnotemark \\ \hline \hline
&& \multicolumn{4}{c|}{CLIP-ViT-B} & \multicolumn{4}{c}{ResNet-50} \\ \hline
& CE 
                        & 80.22 / 80.22 & 27.50 / 27.50 & 65.21 / 61.55 & 72.83 / 56.00
                        & 65.60 / 65.60 & \,1.45 / \,1.37 & 53.21 / 53.21 & 57.08 / 47.25\\
FFT & GCE 
                        & 81.46 / 81.46 & \,6.13 / \,3.54 & 77.63 / 75.87 & \,1.07 / \,1.07
                        & 55.76 / 55.76 & \,5.73 / \,4.72 & 63.17 / 63.18 & \,0.95 / \,0.95 \\
& ELR 
                        & 76.88 / 76.88 & 32.82 / 32.82 & 76.86 / 76.86 & 71.35 / 65.36 
                        & 65.11 / 65.11 & \,8.08 / \,8.08 & 61.18 / 61.18 & 56.88 / 56.88 \\ 
\rowcolor{lightergray}
& CE 
                        & 74.04 / 69.83 & 51.17 / 35.07 & 55.88 / 53.41 & 63.36 / 59.04 
                        & 67.84 / 63.68 & 51.06 / 33.95 & 53.73 / 49.58 & 31.11 / 50.44 \\
\rowcolor{lightergray}
& GCE 
                        & 79.50 / 78.90 & 65.00 / 61.11 & 73.01 / 71.43 & 19.75 / 19.45 
                        & 64.61 / 64.61 & 49.15 / 48.60 & 58.47 / 58.47 & 58.02 / 58.02 \\
\rowcolor{lightergray}
LP & ELR 
                        & 74.19 / 70.85 & 51.50 / 36.71 & 57.66 / 52.28 & 65.89 / 64.46 
                        & 67.50 / 64.44 & 50.79 / 34.72 & 55.56 / 50.92 & 54.45 / 52.91 \\
\rowcolor{lightergray}
& DivideMix 
                        & 79.40 / 79.33 & 69.54 / 69.54 & 76.00 / 76.00 & 70.70 / 70.40 
                        & 70.37 / 70.37 & 56.85 / 56.74 & 64.26 / 64.07 & 60.68 / 60.68 \\
\rowcolor{lightergray}
& UNICON 
                        & 74.44 / 74.44 & 60.63 / 60.63 & 67.75 / 67.71 & 73.54 / 73.54 
                        & 69.03 / 69.03 & 57.89 / 57.89 & 66.69 / 66.69 & 67.95 / 67.95 \\ 

\rowcolor{lightgray}
LP-FFT & Ours
                        & \textbf{83.46 / 83.46} & \textbf{71.27 / 72.17} & \textbf{79.29 / 79.29} & \textbf{79.29 / 79.29} 
                        & \textbf{72.12 / 72.12} & \textbf{58.11 / 58.11} & \textbf{68.33 / 68.33} & \textbf{69.99 / 69.99} \\ 
\thickhline
\end{tabular}
}
\vspace{-15pt}
\end{table*}

\addtocounter{footnote}{1}
\footnotetext[2]{Due to the characteristics of MSN, effficient fine tuning with a small number of epochs is difficult when the number of samples is small. For example, if we run 5 epochs using $10\%$ ($4,750$/$47,500$ samples) clean dataset, it shows $10.14\%$ accuracy. \alg selects about $12.68\%$ of samples at a clean ratio of $91.40\%$ and shows $12.02\%$ accuracy when train 5 epochs. Therefore, we report $\times 5$ times for each epochs. Note that \alg select clean samples every beginning of each epoch, thus we feedforward $\times 0.5$ compared to the other cases.}

\subsection{Classification Result}
\label{sec:exp_result}
In this section, we report the performances of the methods on various PTMs under the CIFAR-100 dataset, Clothing-1M~\cite{xiao2015learning} and WebVision~\cite{li2017webvision}. For CIFAR-100 dataset, we include four noisy label cases \{\textit{Symm} $0.6$, \textit{Symm} $0.9$, \textit{Asym} $0.4$, \textit{Inst} $0.4$\}.

\myparagraph{CIFAR datasets.} The results shown in~\autoref{tab:cifar_synth} demonstrate that \alg, when applied to various PTMs, consistently delivers high performance across different types and severities of noisy labels. Furthermore, the proposed method exhibits the superiority of the FFT and LP tuning mechanisms. Specifically, the ViT-B/16 model consistently outperforms other PTMs.

As discussed in Section~\ref{sec:motivation}, in cases of severe label noise (\ie \textit{Symm} $90\%$ noise), the LP-based approach generally yields more stable performance compared to FFT. Conversely, in cases of less severe label noise (e.g., \textit{Symm} $60\%$ noise), FFT tends to outperform LP. This tendency can be understood by \textbf{Obs 1} in Section~\ref{sec:motivation} that FFT under severe noise can distort the feature extraction. However, \alg approach consistently outperforms both cases. Therefore, it can be concluded that the sequential combination of LP followed by FFT is a valuable way for tuning PTMs on noisy datasets.

\begin{table*}[t]
\centering
\caption{Comparison with LNL algorithms in test accuracy (\%) on Clothing1M and WebVision. We run totally five architectures under the same noisy label setting. The best results are highlited in \textbf{bold}. We report the best/last performance for each experiment. We report both best/last accuracies.}
\vspace{-5pt}
\label{tab:real_world}
\resizebox{1.0\textwidth}{!}{
\begin{tabular}{l|ccccc|ccc|c}
\thickhline
\multirow{3}{*}{Architecture} 
& \multicolumn{9}{c}{Clothing1M} \\ 
& \multicolumn{5}{c|}{LP} & \multicolumn{3}{c|}{FFT} & LP+FFT \\
& CE & GCE & ELR & DivideMix & UNICON & CE & GCE & ELR  & Ours \\ \hline 
ViT-B/16        & 67.83 / 67.54 & 67.46 / 67.46 & 66.91 / 66.91 & 68.13 / 68.13 & 68.42 / 68.42 & 68.98 / 68.98 & 69.74 / 69.74 & 68.73 / 68.73 & \textbf{70.28 / 70.28}\\
ConvNeXt-T      & 64.82 / 64.81 & 64.59 / 64.59 & 64.17 / 64.17 & 66.12 / 65.42 & 67.33 / 66.92 & 68.80 / 68.80 & 68.92 / 68.92 & 69.19 / 68.52 & \textbf{69.63 / 69.63}\\
MAE-ViT-B       & \,5.06 / \,5.06 & \,5.92 / \,5.92 & \,8.28 / \,8.28 & \,8.04 / \,8.04 & \,8.52 / \,8.52 & 61.31 / 61.31 & 60.80 / 60.80 & 61.51 / 61.51 & \textbf{61.96 / 61.96}\\
MSN-ViT-B       & \,6.77 / \,6.77 & \,6.20 / \,6.20 & \,7.64 / \,7.64 & \,6.42 / \,6.42 & \,6.31 / \,6.31 & 66.88 / 63.38 & 67.06 / 65.41 & 66.32 / 66.32 & \textbf{69.13 / 69.13} \\
ResNet-50          &  \,7.08 / \,7.08 & \,7.18 / \,7.18 & \,6.68 / \,6.68 & \,8.13 / \,8.13 & \,8.24 / \,8.24 & 66.10 / 66.02 & 66.19 / 66.19 & 66.19 / 66.19 & \textbf{66.31 / 66.31}\\
\hline \hline 
\multirow{3}{*}{Architecture} 
& \multicolumn{9}{c}{WebVision} \\
& \multicolumn{5}{c|}{LP} & \multicolumn{3}{c|}{FFT} & LP+FFT \\
& CE & GCE & ELR & DivideMix & UNICON & CE & GCE & ELR  & Ours \\ \hline 
ViT-B/16        & 84.62 / 84.48 & 84.32 / 84.24 & 84.48 / 84.32 & 84.72 / 84.72 & 85.68 / 85.68 & 84.20 / 83.04 & 83.40 / 83.40 & 84.92 / 83.72 & \textbf{85.96 / 85.92} \\
ConvNeXt-T      & 85.24 / 85.24 & 85.12 / 85.04 & 86.28 / 86.28 & 86.40 / 86.40 & 86.24 / 86.24 & 84.00 / 82.68 & 85.40 / 84.92 & 84.52 / 83.44 & \textbf{87.16 / 86.44} \\
MAE-ViT-B       & 48.00 / 48.00 & 47.32 / 47.28 & 49.76 / 49.76 & 59.40 / 58.44 & 56.96 / 53.80 & 67.48 / 65.64 & 63.16 / 62.84 & 67.80 / 67.80 & \textbf{69.45 / 68.45} \\
MSN-ViT-B       & 77.40 / 77.40 & 74.40 / 74.40 & 74.00 / 74.00 & 76.56 / 76.40 & 77.72 / 77.34 & 77.04 / 77.80 & 72.28 / 72.28 & 74.88 / 72.28 & \textbf{78.36 / 75.40} \\
ResNet-50       & 84.88 / 84.72 & 81.68 / 81.68 & 84.96 / 84.96 & 85.16 / 85.16 & 85.04 / 85.04 & 78.00 / 76.44 & 77.04 / 70.92 & 80.44 / 77.44 & \textbf{85.36 / 85.36} \\
\thickhline
\end{tabular}}
\vspace{-15pt}
\end{table*}

\myparagraph{Real-world tasks.} We evaluate the performance of \alg on larger datasets, namely Clothing1M and WebVision. The results in Table~\ref{tab:real_world} consistently demonstrate that \alg improves performance on these datasets. This highlights the ability of \alg to handle noisy labels in real-world scenarios. However, some models, like MAE, MSN, and ResNet on Clothing1M, show a drop in performance due to the fine-grained nature compared to the training dataset (\ie ImageNet). Nonetheless, our algorithm performs better by utilizing FFT in the second step.

\subsection{Analysis}
\label{sec:exp_analysis}
We design our analyses to answer the following key questions: (1) Is \alg efficient compared to previous algorithms? (2) Is \alg sensitive to its hyperparameters? (3) Does \alg perform well on larger models? (4) Which components of \alg contribute to its improved performance?  By exploring these questions, our analyses provide additional insights and explanations to understand \alg.
Additional analyses including cleansing performance are described in Appendix~\ref{app:add}.

\begin{figure*}[t]
    \centering
    \begin{subfigure}{0.235\textwidth}
            \includegraphics[width=1.\linewidth]{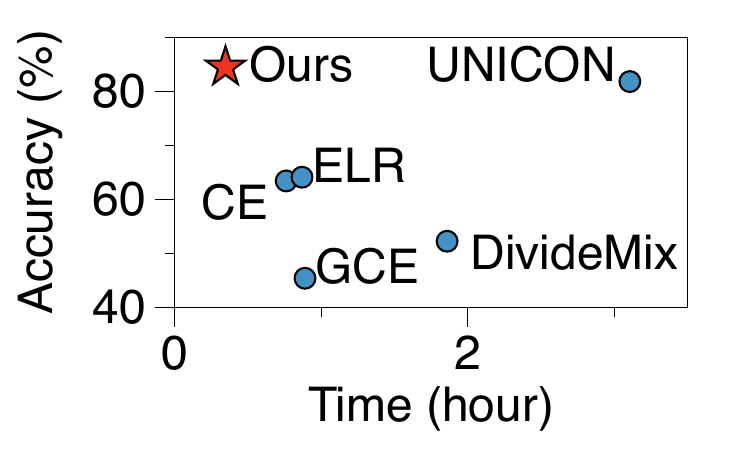}
            \vspace{-15pt}
            \caption{Time}
            \label{fig:time}
    \end{subfigure}
    \hfill
    \begin{subfigure}{0.235\textwidth}
            \includegraphics[width=1.\linewidth]{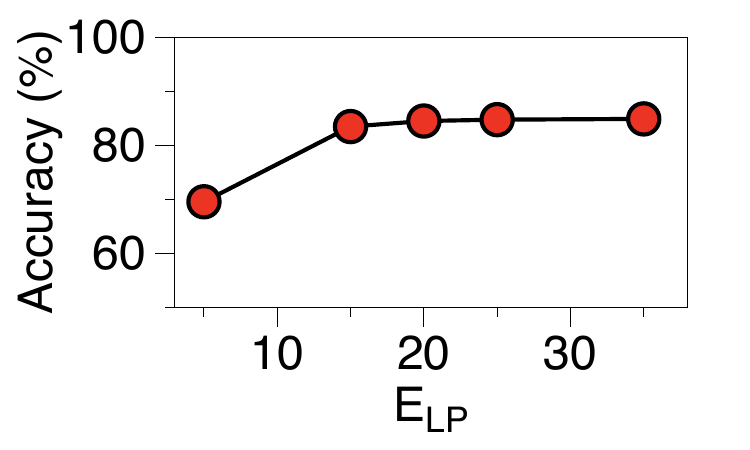}
            \vspace{-15pt}
            \caption{$E_\text{LP}$ Sensitivity}
            \label{fig:lp_epoch}
    \end{subfigure}
    \hfill
    \begin{subfigure}{0.235\textwidth}
            \includegraphics[width=1.\linewidth]{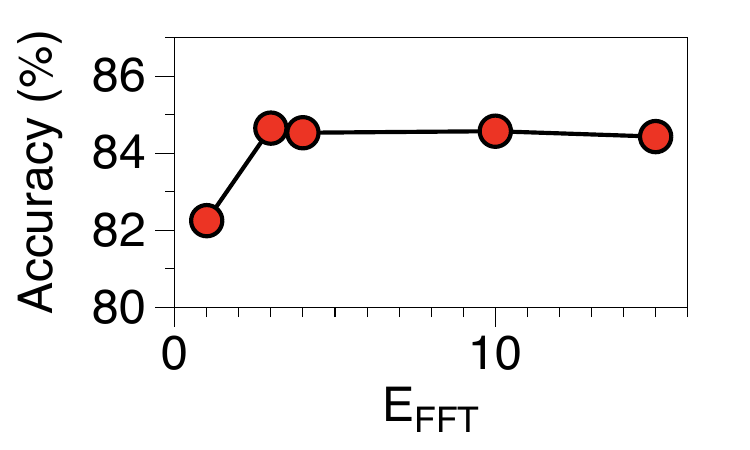}
            \vspace{-15pt}
            \caption{$E_\text{FFT}$ Sensitivity}
            \label{fig:fft_epoch}
    \end{subfigure}
    \hfill
    \begin{subfigure}{0.235\textwidth}
            \includegraphics[width=1.\linewidth]{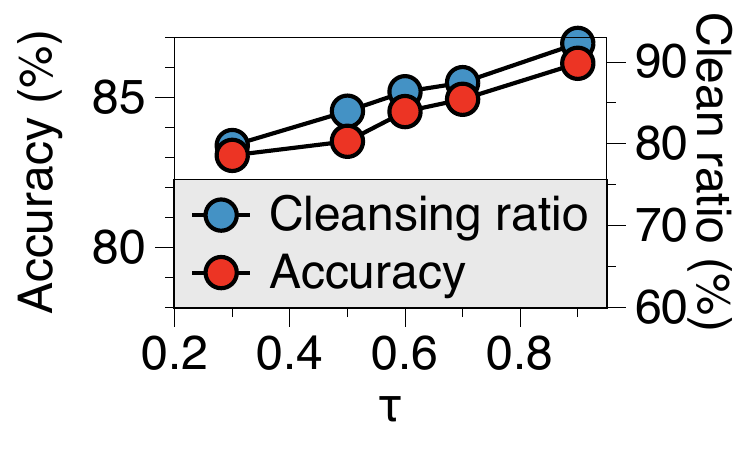}
            \vspace{-15pt}
            \caption{GMM $\tau$ Sensitivity}
            \label{fig:gmm}
    \end{subfigure}
    \vspace{-5pt}
    \caption{In-depth analysis of the proposed algorithm. Training time, parameter sensitivity, respectively. We run CIFAR-100 with $90\%$ symmetric noise.}
    \label{fig:out_domain}
\end{figure*}

\myparagraph{Training time.} To assess the efficiency of the proposed algorithm, we examine the correlation between training time and test accuracy as depicted in~\autoref{fig:time}. We \textbf{\textit{run FFT}} on the CIFAR-100 benchmark under $90\%$ noisy ratio. We run each algorithm for $5$ epochs. As described in~\autoref{fig:time}, \alg shows better performance ($+2.69\%$ compared to UNICON) while spending the smallest training time (\ie $\times 0.12$ compared to UNICON). This is because leveraging noisy labels by giving pseudo-labels for them, such as DivideMix and UNICON, requires a significant amount of time due to their computationally expensive nature. Furthermore, robust loss-based algorithms exhibit reduced computational costs compared to the pseudo-label based methods but they used entire training samples, while \alg utilize part of a clearer training dataset. This analysis emphasizes the efficiency of \alg.

\myparagraph{Hyperparameter sensitivitiy.}
In the proposed method, we analyze three types of hyperparameters: (1) the number of epochs for LP denoted as $E_{\text{LP}}$, (2) the number of epochs for FFT denoted as $E_{\text{FFT}}$, and (3) the GMM threshold $\tau$. Our experiments on CIFAR-100 with $90\%$ symmetric noise reveal that increasing the number of epochs for LP improves performance until reaching $20$ epochs, beyond which the gain becomes negligible. Similarly increasing the number of FFT epochs up to $5$ provides similar performance due to the similar sample size per epoch. Moreover, increasing the GMM threshold results in a correct subset of improved performance. Therefore, careful tuning of these hyperparameters can contribute to enhanced performance.

\myparagraph{Larger model.} We also examine the performance of the proposed algorithm using larger models (ViT-L/16 and CLIP-ViT-L/14) on the CIFAR-100 dataset with $90\%$ symmetric case. As shown in~\autoref{tab:large}, the proposed algorithm consistently outperforms the other FFT and LP cases. Notably, when using larger models, the performance tends to improve compared to the results in Table~\ref{tab:cifar_synth}. However, it is important to consider that larger models require more computation resources and time for tuning due to the increased number of model parameters.

\begin{wraptable}[5]{r}{0.4\textwidth}
\vspace{-15pt}
\caption{Component analysis.}
\label{tab:component}
\vspace{-15pt}
\resizebox{0.4\textwidth}{!}{
\begin{tabular}{ccc|cc}\\\thickhline
LP & Cleansing & FFT & ViT-B/16 & CLIP-ViT-B \\ \hline
    &   Multiple & \cmark & 45.51 & 30.07 \\
\cmark &  None & \cmark & 45.90 & 47.47 \\
\cmark & Once & \cmark & 83.00 & 60.80  \\
\cmark & Multiple & \cmark & 83.33 & 72.17 \\ 
\thickhline
\end{tabular}}
\end{wraptable}


\myparagraph{Component analysis.} 
To evaluate the impact of each component of \alg, we conducted experiments on the CIFAR-100 dataset under $90\%$ symmetric case. Specifically, we examined three components: (1) LP, (2) Cleansing, and (3) FFT. Regarding Cleansing, we explored different types: None (no cleansing), Once (cleansing once right before FFT), and Multiple (cleansing at the beginning of Step 2). The results, as shown in~\autoref{tab:component}, indicate that the configuration involving all components achieves the best performance. Additionally, LP acts as a buffer to prevent the feature extractor from being distorted, as evidenced by the performance drop observed in the first row. When cleansing is omitted, a significant decrease in performance is also observed. 

\begin{table*}[t]
\centering
\vspace{-5pt}
\caption{Analysis on larger PTMs. We use large models of ViT and CLIP, \ie ViT-L/16 and CLIP-ViT-L/14, respectively. We run CIFAR-100 with $90\%$ symmetric noise.} 
\label{tab:large}
\vspace{-5pt}
\resizebox{0.98\textwidth}{!}{
\begin{tabular}{c|ccccc|ccc|c}
\thickhline
\multirow{2}{*}{Method}
                        & \multicolumn{5}{c|}{Linear Probing} 
                        & \multicolumn{3}{c|}{Full Fine-Tuning} 
                        & \multicolumn{1}{c}{LP + FFT} \\ 
                        
                            & CE & GCE & ELR & DMix & UNICON & CE & GCE & ELR & Ours \\\hline \hline
ViT-L/16
                        & 81.20 / 60.77 & 70.02 / 70.02 & 81.70 / 79.46 & 84.58 / 84.58 & 84.87 / 84.87 
                        
                        & 84.37 / 53.08 & 42.71 / 42.71 & 8022 / 49.78 & \textbf{87.02 / 85.90} \\

CLIP-ViT-L/14
                        & 41.24 / 41.24 & 36.48 / 36.48 & 41.01 / 41.01 & 80.68 / 80.13 & 75.83 / 75.83
                        & 63.30 / 38.59 & 71.93 / 71.70 & 68.16 / 48.27 & \textbf{81.17 / 81.17 }\\

\thickhline
\end{tabular}
}
\vspace{-10pt}
\end{table*}

\section{Related Work}
\label{sec:related}

\myparagraph{Learning with Noisy Labels.} Noisy label problem has been explored extensively in recent researches~\cite{zhang2017understanding, han2018co, li2020dividemix, cheng2021learning, kim2021fine, xia2022sample, karim2022unicon, cheng2023mitigating}. The authors of  \cite{zhang2017understanding} showed that DNNs easily memorize randomly labeled training data, numerous studies have addressed the memorization problem under label noise. Existing methods mainly address this problem by (1) detecting corrupted instances and only using label information of clean examples~\cite{han2018co, yu2019does, li2020dividemix, cheng2021learning, kim2021fine, xia2022sample} (2) designing loss functions~\cite{zhang2018generalized, wang2019symmetric, zhou2021asymmetric} or regularization terms~\cite{liu2020early, ko2022alasca, cheng2023mitigating} with robust behaviors.

Recently, the majority of the research~\cite{zheltonozhskii2022contrast, karim2022unicon, li2022selective} is focused on applying self-supervised approaches to construct robust feature extractors on label noise. The authors of~\cite{zheltonozhskii2022contrast} proposed C2D to run semi-supervised approaches~\cite{li2020dividemix, liu2020early} with the initial parameters from the SimCLR~\cite{chen2020simple} and showed significant performances. However, these approaches may be over-complicated requiring hyperparameter tuning for different datasets, as well as significant computation resources.

\myparagraph{Pre-trained Models.} Recently, several studies have demonstrated that PTMs, which are trained on the large image\,\cite{ridnik2021imagenetk} or text corpora, can learn universal visual or language representations that are useful for downstream computer vision or natural language processing tasks. This has eliminated the need to train a new model from scratch. With the advancement of computational power and development of deep models such as GPT-3\,\cite{brown2020language}, Vision Transformer\,\cite{dosovitskiy2021an}, and ConvNext\,\cite{liu2022convnet}, the capabilities of PTMs have greatly improved. Utilizing PTMs has been considered as an effective solution for multi-modal models such as CLIP\,\cite{radford2021learning} and Data2Vec\,\cite{baevski2022data2vec}, which can effectively represent various types of domains.

As researchers make pre-trained weights of PTMs available to the open-source community, there is growing interest in finding ways to effectively use these pre-trained weights. For example, a lot of recent research is being focused on utilizing large pre-trained models for prompt learning\,\cite{zhang2022differentiable} or in-context learning\,\cite{liu-etal-2022-makes}, with the goal of achieving good results in few-shot learning scenarios. However, relatively only a few studies have explored using these PTMs in a robust learning framework to handle label noise. Recently, \cite{zhu2022detecting} suggested the applying self-supervised PTMs\,\cite{chen2020simple} or large PTM\,\cite{radford2021learning} for detecting corrupted labels without training, however, these methods applied the KNN technique, which requires heavy computational consumption~\cite{li2022selective}.

\section{Conclusion}
\label{sec:conclusion}

This study presents a new algorithm known as \alg, which aims to address the problem of noisy labels by effectively utilizing large pre-trained models. The algorithm has two primary objectives: (1) robustly utilizing large pre-trained models and (2) effectively reducing the impact of noisy labels. To achieve these goals, the proposed algorithm employs linear probing and then full fine-tuning on a cleansed subset of the original training dataset. The improved performance of the proposed algorithm is demonstrated through extensive experiments on synthetic CIFAR-100 dataset, as well as real-world benchmarks such as Clothing 1M and WebVision on various types of PTMs. Moreover, the algorithm also shows not only performance improvements but also reduced computational costs.

\bibliographystyle{abbrv}
\bibliography{ref}

\clearpage
\appendix
\onecolumn
\begin{center}
    \noindent\rule{0.99\textwidth}{1.0pt}\\
    \vspace{0.12in}
    \textbf{\Large{- Supplementary -\\}}
    \textbf{\Large{\mytitle}}\\
    \vspace{0.12in}
    \noindent\rule{0.99\textwidth}{1.0pt}
\end{center}

\section{Implementation Details}
\label{app:imp}

\subsection{Dataset Description}\label{app:dataset}

\begin{table}[h]
\centering
\caption{Description of a dataset.}
\label{tab:dataset}
\vspace{3pt}
\begin{tabular}{c|c|c|c|c}
\thickhline

Dataset         & \# class      & \# train      & \# valid      & \# test   \\ \hline
CIFAR-100       & 100           & 47.5K        & 2.5K          & 10K      \\ \hline
Clothing 1M     & 14            & 1M            & 14K           & 10K        \\ \hline
WebVision       & 1,000         & 2.4M          & -             & 50K        \\ \hline

\thickhline
\end{tabular}
\end{table}

Within this section, we provide an overview of the datasets and statistics employed in our evaluation. We conducted assessments on three distinct datasets, namely CIFAR-100, Clothing 1M, and Webvision. These datasets were subjected to diverse configurations of noisy labels in order to analyze their impact on the performance of the model.

\myparagraph{CIFAR-100~\cite{krizhevsky2009learning}.} The CIFAR-100 dataset comprises a collection of $60,000$ $32\times32$ color images, which are classified into $100$ distinct classes. The dataset is split into a training set containing 50,000 images ($47,500$ for training and $2,500$ for validation) and a testing set containing $10,000$ images. These $100$ classes are further grouped into $20$ superclasses, with each superclass containing multiple classes. For the training set, there are $500$ images per class, while the testing set includes $100$ images per class. Notably, each image in the CIFAR-100 dataset is associated with both a "fine" label that corresponds to the specific class and a "coarse" label that corresponds to the superclass it belongs to. We leveraged fine labels for this paper.

\myparagraph{Clothing 1M~\cite{xiao2015learning}.} Clothing 1M is a dataset comprising a substantial collection of 1 million clothing images distributed across 14 distinct classes. The dataset is known to possess a noisy label environment, as it was assembled from multiple online shopping websites, resulting in a considerable number of mislabeled samples. According to a study by Song et al. (2022)~\cite{song2022learning}, Clothing 1M exhibits an estimated noise level of $38.5\%$. Moreover, within this dataset, there are subsets of $50,000$, $14,000$, and $10,000$ correctly labeled subsets allocated for training, validation, and testing, respectively. 

\myparagraph{WebVision~\cite{li2017webvision}.} The WebVision dataset serves as a valuable resource for exploring the field of learning visual representations from web data, particularly in the presence of noise. It is an extensive dataset consisting of over 2.4 million images obtained by crawling the Flickr website and conducting searches on Google Images. The dataset's large-scale nature and diverse image sources make it well-suited for studying and developing algorithms that can effectively handle the challenges posed by web data.

\subsection{Data Preprocessing}
\subsubsection{Noisy Label Generation for Synthetic Datasets}

To introduce label noise into the datasets, we employ a synthetic approach based on the methodology proposed by Liu et al. (2020)~\cite{liu2020early}. We generate both symmetric and asymmetric noise configurations using this method. For symmetric noise, we randomly select a portion of the labels and modify them using a uniform random process. This alteration introduces randomness and noise into the selected labels. Regarding asymmetric noise specifically applied to the CIFAR-100 dataset, we partition the dataset into 20 super-classes, each containing five classes. Within each super-class, we shift the labels of each class to the subsequent class, creating a systematic deviation from the original labeling scheme. Furthermore, we incorporate instance-based noise by employing the approach outlined by Cheng et al. (2021)~\cite{cheng2021learning}. This method allows us to generate noise patterns unique to each individual instance. To compare its performance against existing label noise learning methods, we sample flip rates from a truncated normal distribution. The mean of the distribution corresponds to the specified noise ratio, with a standard deviation of $0.1$. The flip rates are sampled within the range of $0$ to $1$. To parameterize the instance-dependent label noise, we sample the noise parameters, denoted as $W_I$, from a standard normal distribution. The size of $W_I$ is determined by the length of each feature ($L$) multiplied by the number of classes ($C$), resulting in a matrix of size $L \times C$. This process ensures the generation of diverse and instance-specific label noise patterns.

\subsubsection{Data Augmentation} 
All image preprocessing is done using the officially released preprocessor that comes with the corresponding PTMs (ViT, CLIP, MAE, MSN, ResNet and ConvNeXt). In addition, we perform image augmentation to generate similar images in order to implement instance-centric consistency. The additional image augmentation for each dataset is as follows: Random Horizontal Flip $\rightarrow$ Random Affine (CIFAR-100) and Random Resized Crop $\rightarrow$ Random Horizontal Flip $\rightarrow$ Color Jitter (Clothing1M, WebVision).

\subsection{Experiments Setting}

To conduct a hyperparameter search, we established the search space as outlined in Table~\ref{tab:search}. This exploration process was carried out specifically on the CIFAR-100 dataset, with the presence of symmetric noise accounting for $90\%$ of the labels. During training, all models utilized cross-entropy loss as their objective function. For a comprehensive understanding of the specific hyperparameters employed, please refer to Table~\ref{tab:hyp}.

In the case of DivideMix and UNICON, a halved batch size was utilized. This adjustment was necessary as these models incorporate two data loaders simultaneously, requiring a modified batch size configuration.

\begin{table}[h]
\centering
\caption{Hyperparameter searching space.}
\label{tab:search}
\vspace{3pt}
\begin{tabular}{c|c}
\thickhline
Hyperparameter      & Search Space \\ \hline
Learning Rage (LR)  & \{$10^{-1}$, $10^{-2}$, $10^{-3}$, $10^{-4}$, $10^{-5}$\} \\ \hline
Optimizer (LP)      & \{sgd, adamw\} \\ \hline
Weight Decay (WD)   & \{$10^{-2}$, $10^{-3}$, $10^{-4}$, $10^{-5}$ $10^{-6}$\} \\ \hline
Batch size          & \{$32$, $64$, $128$\} \\ \hline
\thickhline
\end{tabular}
\end{table}

\begin{table}[h]
\centering
\caption{Hyperparameters for experiments.}
\label{tab:hyp}
\vspace{3pt}
\begin{tabular}{c|c|c|c|c|c|c}
\thickhline
Model              & ViT                & ConvNeXt              & MAE                   & MSN                   & CLIP                  & ResNet-50 \\ \hline
Optimizer (LP)     & sgd                & sgd                   & adamw                 & sgd                   & adamw                 & adamw     \\ \hline
Learning rate (LP) & {$1\times10^{-2}$} & {$1\times10^{-1}$}    & {$1\times10^{-2}$}    & {$1\times10^{-2}$}    & {$1\times10^{-3}$}    & {$1\times10^{-3}$}  \\ \hline
Weight Decay (LP)  & {$1\times10^{-3}$} & {$1\times10^{-4}$}    & {$1\times10^{-5}$}    & {$1\times10^{-2}$}    & {$1\times10^{-6}$}    & {$1\times10^{-5}$}  \\ \hline
Optimizer          & sgd                & adamw                   & adamw                 & sgd                   & adamw                 & adamw     \\ \hline
Learning rate      & {$1\times10^{-2}$} & {$1\times10^{-4}$}    & {$1\times10^{-4}$}    & {$1\times10^{-3}$}    & {$1\times10^{-5}$}    & {$1\times10^{-3}$}  \\ \hline
Weight Decay       & {$1\times10^{-5}$} & {$1\times10^{-4}$}    & {$1\times10^{-3}$}    & {$1\times10^{-5}$}    & {$1\times10^{-5}$}    & {$1\times10^{-5}$}  \\ \hline
Batch size         & \multicolumn{6}{|c}{$128$}  \\ \hline
\thickhline
\end{tabular}
\end{table}

\subsection{Experiment Environment}
During the training process involving FFT, DivideMix, and UNICON models, we employed a setup consisting of four V100 GPUs for DivideMix and UNICON. Other methods were adequately trained using two RTX 3090 GPUs. However, it's important to note that CLIP, due to its resource requirements, necessitated a minimum of three A5000 GPUs to facilitate training.

To expedite the experimental process, we made use of a combination of hardware resources, including 12 RTX 3090 GPUs, 6 A5000 GPUs, and 4 V100 GPUs. These GPUs were utilized interchangeably, ensuring efficient utilization of computational resources throughout the experiments.
\section{Baseline}
\label{app:base}

\subsection{Pre-trained Model Description}

In this section, we introduce the PTMs that we mainly use to implement our proposed method, \alg. All pre-trained weights are officially released in HuggingFace~\cite{wolf-etal-2020-transformers} and we are easily accessible to these models. 

\myparagraph{Vision Transformer (ViT~\cite{dosovitskiy2021an}).} The Vision Transformer utilizes a Transformer-inspired architecture to classify images by dividing them into fixed-size patches, linear embedding each patch and incorporating position embeddings. These resulting vectors are then processed by a standard Transformer encoder and a learnable "classification token" is added to the sequence for classification. We mainly use ViT-B/16 which has 86M parameters and is pre-trained on ImageNet-1K. For the large model analysis, we use ViT-L/16 which is pre-trained on ImageNet-21K.

\myparagraph{Contrastive Language-Image Pre-training (CLIP~\cite{radford2021learning}).} CLIP is a highly efficient method for learning image representations through natural language supervision. It employs a simplified version of ConVIRT~\cite{zhang2020contrastive}, and utilizes a joint training approach for both an image encoder and a text encoder, with the goal of correctly predicting the pairing of a batch of (image, text) training examples. In our experiment, we utilize CLIP-ViT-B which applies CLIP pretraining method to ViT architecture. 

\myparagraph{Residual Nets (ResNet-50~\cite{he2016deep})} ResNet is a widely adopted deep neural network architecture that has gained popularity in diverse computer vision tasks. It introduces the concept of residual blocks, where the input of a layer is added to its output, simplifying optimization and enhancing performance. Through the stacking of multiple layers with residual blocks, ResNet enables the creation of a sophisticated model capable of capturing intricate patterns and understanding complex relationships within the data. In our study, we employ ResNet-50 as the primary architecture, which contains 25.6 million parameters.

\myparagraph{Convolutional Neural Networks (ConvNeXt~\cite{liu2022convnet}).} ConvNeXt is a transformation of standard ResNet into the design of a Vision Transformer through the gradual process of modernization by uncovering several crucial components that contribute to the performance difference. This exploration resulted in a family of pure ConvNet models, known as ConvNeXt. These models are entirely built from standard ConvNet modules and have been found to be competitive in terms of accuracy and scalability when compared to Transformers. We mainly use ConvNeXt-T/224 which is pre-trained on ImageNet-1K, and consists of 28M parameters. 

\myparagraph{Masked Autoencoders (MAE~\cite{he2022masked}).} 
Masked Auto Encoder (MAE) is a self-supervised learning framework designed for training large pre-trained models. It comprises an encoder and a decoder component. During training, the encoder and decoder are utilized to extract features and reconstruct images based on that features which comes from the input images with partially removed patches. This approach encourages the model to learn meaningful representations from incomplete data. In our study, we utilize the MAE-ViT-B model, which is based on the Vision Transformer (ViT) architecture. This model possesses $86M$ parameters and has been trained on the ImageNet-1K dataset.

\myparagraph{Masked Siamese Networks (MSN~\cite{assran2022masked}).} 
Masked Siamese Network (MSN) serves as a self-supervised learning framework employed for training large pre-trained models. Its objective is to minimize the feature distance between the representation of the original image and the representation of the masked image. In our study, we utilize the MSN-ViT-B model, which consists of $86M$ trainable parameters and has undergone training on the ImageNet-1K dataset.

\subsection{Learning with Nosiy Labels}

\myparagraph{Generalized Cross Entropy (GCE~\cite{zhang2018generalized}).}
Generalized Cross Entropy (GCE) is a denoising mechanism that utilizes a robust-loss approach. The loss function is defined as:
\begin{equation*}
\mathcal{L}_{\text{GCE}}(f(x), y) = \frac{(1-f(x)[y]^q)}{q},
\end{equation*}
In the equation, $f(x)$ represents the softmax output, and $f(x)[y]$ denotes the $y$-th value of the output. The hyperparameter $q \in (0,1]$ controls the loss. As $q$ approaches 0, it resembles the conventional cross entropy loss, while $q = 1$ is equivalent to the Mean Absolute Error loss.

By adjusting the value of $q$, the GCE loss can prioritize or ignore difficult-to-learn samples. When $q$ is larger, it tends to disregard challenging samples, whereas smaller values of $q$ pay more attention to such samples. In our study, we utilize $q=0.7$, as proposed by the original paper. This choice allows the loss function to effectively handle difficult-to-learn samples, particularly those associated with noisy labels.

\myparagraph{Early Learning Regularizer (ELR)~\cite{liu2020early}.} Early Learning Regularizer (ELR) is a type of regularizer that incorporates predicted labels using a moving average. It is based on the observation that initially, the trained model cannot memorize noisy labels, but gradually it do so as training progresses. The original paper proposes two algorithms: ELR and ELR+. In our study, we utilize only the ELR regularizer, as reducing computational cost is a priority when leveraging large-scale pre-trained models.

The ELR loss function is defined as follows:
\begin{equation*}\label{eq:elr}
\mathcal{L}{\text{ELR}}(f(x), y) = \mathcal{L}{\text{CE}}(f(x), y) + \lambda \log (1- \langle \mathbf{p}, \mathbf{t} \rangle),
\end{equation*}
Here, $\langle \mathbf{p}, \mathbf{t} \rangle$ represents the inner product between the softmax output of the model, denoted as $\mathbf{p} = \text{Softmax}(f(\mathbf{x}))$, and the moving average (MA) value of the model output. The MA value is calculated as $\mathbf{t} \leftarrow \beta \mathbf{t} + (1-\beta) \mathbf{p}$, where $\beta$ is the MA parameter. The regularization term of the ELR loss function is weighted by $\lambda$. It is important to note that we follow the hyperparameters defined in the original paper.

\myparagraph{DivideMix~\cite{li2020dividemix}.} DivideMix is a denoising method that involves training two networks concurrently. One network divides the input samples into two groups: those considered as having clean labels and those labeled with pseudo-labels since they are regarded as noiy labels. This division is accomplished using a Gaussian Mixture Model (GMM), which shares similarities with our split mechanism. The other network is trained using augmented samples from the subset considered as clean. In this process, MixUp augmentation is employed, generating convex combinations between samples from the clean subset and the noisy subset. It is important to note that DivideMix requires a significant amount of time due to multiple feed-forward steps involved in verifying the noisy labels, generating pseudo-labels, and extracting features from the MixUped images.

\myparagraph{UNICON~\cite{karim2022unicon}.} UNICON is an enhanced version of the DivideMix method, designed to address the class imbalance issue present in the clean subset. It recognizes that the clean dataset in DivideMix may suffer from a lack of balance among different classes. To overcome this, UNICON introduces a uniformly subsampling approach for the clean dataset, ensuring a more equitable representation of classes during training. Additionally, UNICON incorporates contrastive loss to enhance the feature extractor's capabilities. However, it is worth noting that the utilization of contrastive loss in UNICON leads to a significant increase in computational cost due to the additional feed-forwards required to compute this loss function.
\section{Additional Experiments}
\label{app:add}

\subsection{Cleansing Performance Anlaysis}

\begin{table}[ht]
\caption{The cleansing performance varying the GMM threshold.}
\label{tab:purity}
\centering
\resizebox{0.8\textwidth}{!}{
\begin{tabular}{c|ccccc}
\thickhline
& \multicolumn{5}{c}{Symmetric 0.9} \\ \hline
GMM $\tau$& 0.3 & 0.5 & 0.6 & 0.7 & 0.9  \\
\# Sample & 3500 & 3600 & 3600 & 3600 & 3600 \\
Purity & 78.02 & 80.41 & 83.47 & 84.33 & 88.63\\ 
Acc     & 82.49/80.12 & 82.60/82.21 & 84.22/82.86   & 84.93/83.99  & 85.25/85.25 \\ \hline
& \multicolumn{5}{c}{Symmetric 0.6} \\ \hline
GMM $\tau$& 0.3 & 0.5 & 0.6 & 0.7 & 0.9  \\
\# Sample & 16700 & 16500 & 16300 & 16100 & 15200 \\
Purity & 98.85 & 99.00 & 98.98 & 99.16 & 99.26\\
Acc     & 90.70/90.70   & 90.65/90.65   & 90.52/90.91   & 90.63/90.86   & 90.68/90.68 \\
\thickhline
\end{tabular}}
\end{table}

We evaluate the effectiveness of cleansing performance on CIFAR-100 dataset, considering different levels of symmetric noise at $90\%$ and $60\%$, while varying the Gaussian Mixture Model (GMM) threshold $\tau$. For the data with $90\%$ noise, we observe that increasing the threshold leads to improved purity of the cleansed data, accompanied by higher accuracy. On the other hand, in the case of $60\%$ noise data, we find that even with a low threshold of $0.3$, the purity of the cleansed data remains high.


\end{document}